\documentclass[cameraready]{Interspeech}

\title{From Tokens to Faces: Investigating Discrete Speech Representations for 3D Facial Animation}

\author[affiliation={1,2}, orcid=0000-0002-3411-0959,correspondingauthor]{Pedro R.}{Correa}
\author[affiliation={2}, orcid=0000-0002-9909-6078]{Olivier}{Perrotin}
\author[affiliation={3},orcid=0009-0007-5956-4133]{Samir}{Sadok}
\author[affiliation={1}, orcid=0000-0002-1534-5744]{\\Paula D. P.}{Costa}
\author[affiliation={2}, orcid=0000-0002-8296-5177, correspondingauthor]{Thomas}{Hueber}

\address{
    $^1$ Univ. Estadual de Campinas (UNICAMP), Brazil\\
    $^2$ Univ. Grenoble Alpes, CNRS, Grenoble INP, GIPSA-lab, France \\
    $^3$ Inria at Univ. Grenoble Alpes, CNRS, LJK, France
}

\email{p243236@dac.unicamp.br, olivier.perrotin@grenoble-inp.fr, Samir.sadok@inria.fr, paulad@unicamp.br, thomas.hueber@grenoble-inp.fr}


\usepackage{mathabx}
\definecolor{gipsaPurple}{rgb}{0.34, 0.18, 0.35}
\definecolor{gipsaBlue}{rgb}{0, 0.34, 0.61}
\definecolor{gipsaDark}{rgb}{0.098, 0.133, 0.220}
\definecolor{gipsaDPC}{rgb}{0.64, 0.70, 0.14}
\definecolor{gipsaDAUTO}{rgb}{0.68, 0.20, 0.38}
\definecolor{gipsaPAD}{rgb}{0.745, 0.0, 0.0}
\definecolor{gipsaPSD}{rgb}{0.922, 0.49, 0.137}
\definecolor{gipsaPPC}{rgb}{0.0, 0.686, 0.314}
\definecolor{gipsaGAIA}{rgb}{0.0, 0.686, 0.941}

\newcommand{\symbsigcircle}{\large$\color{gipsaPSD}{\bullet}$\scriptsize}

\newcommand{\symbsigsquare}{\large$\color{gipsaGAIA}{\sqbullet}$\scriptsize}
\newcommand{\symbsigdiamond}{\large$\color{gipsaPPC}{\blackdiamond}$\scriptsize}
\newcommand{\symbsigblank}{\large$\color{white}{\bullet}$\scriptsize}

\newcommand{\B}{$\mathsf{Base}$\xspace}
\newcommand{\Vsota}{$\mathsf{V_{SOTA}}$\xspace}
\newcommand{\Vour}{$\mathsf{V_{OURS}}$\xspace}

\newcommand{\Hp}{$\hat{H}(\mathcal{P}|t)$\xspace}
\newcommand{\Hv}{$\hat{H}(\mathcal{V}|t)$\xspace}
\newcommand{\R}{$R^2$\xspace}

\newcommand{\GRU}{{GRU}}
\newcommand{\T}{{T.}}
\newcommand{\HuBERT}{{HB}}
\newcommand{\CosyVoice}{{CV2}}
\newcommand{\SpeechTokenizer}{{ST}}
\newcommand{\WavTokenizer}{{WT}}

\usepackage{comment}
\usepackage{multirow}
\usepackage{tipa}

\begin{document}

\maketitle


\begin{abstract}

     The choice of speech representation is critical in speech-driven 3D facial animation. Representations differ in what they encode: SSL features emphasize segmental and semantic cues, neural codecs yield latents optimized for acoustic reconstruction, and ASR-style objectives produce label-based spaces. We evaluate four speech representation families for 3D facial synthesis, comparing their facial reconstruction quality across two facial decoders using objective metrics and a perceptual evaluation. We additionally conduct probing analyses that relate tokenized representations to phonetic units and to articulatory deformations. We found that encoding phonetic classes is beneficial for accurate facial animation prediction on both semantic and label-based representations with comparable facial animation quality. From the latter, we introduce an Audio Visual Text-to-Speech (AVTTS) pipeline that leverages, as a shared space, discrete representations to decode speech and 3D facial motion. 
\end{abstract}

\section{Introduction}

\begin{figure}[tbp]
    \centering
    \includegraphics[width=1.0\linewidth]{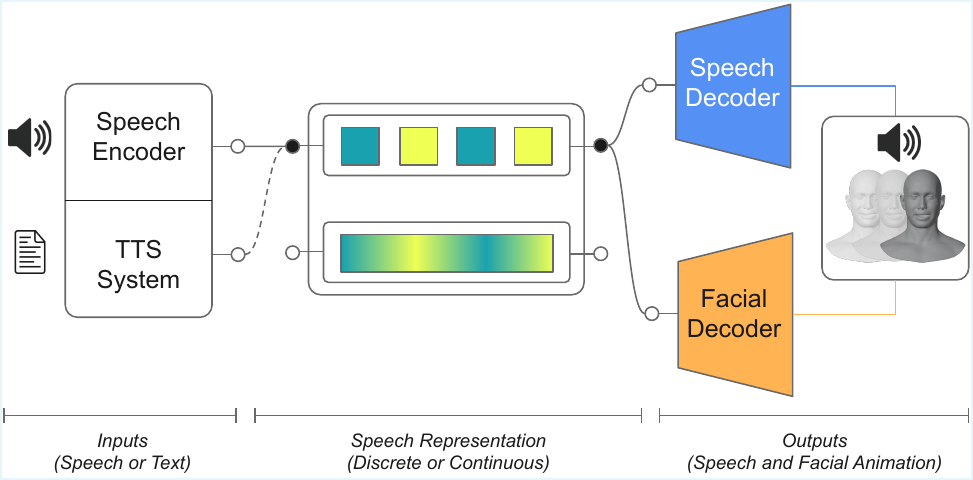}
    \caption{Speech representations (middle) are extracted either from input speech using a speech encoder (top) or produced within a text-to-speech (TTS) pathway conditioned on text and reference speech (bottom). Both discrete and continuous representations are illustrated, with colour coding indicating emphasis on acoustic vs. semantic information. The shared representation can then be decoded into speech audio (speech decoder) and 3D facial motion (facial decoder).}
    \label{fig:general_pipeline}
    \vspace{-1.5em}
\end{figure}

Speech-driven 3D facial animation aims to synthesize temporally coherent and accurate facial movements directly from speech signals~\cite{Cohen1993, Pelachaud1996, karras2017audio, richard2021meshtalk}. Central to this task, most recent architectures (Fig.~\ref{fig:general_pipeline}) use a speech representation bottleneck, which may capture discrete phonetic class information, continuous articulatory dynamics, and prosody at the same time. While recent systems achieve good results, it remains unclear which properties and information in speech representations are truly necessary to drive accurate and natural facial animation.

Most state-of-the-art (SOTA) facial animation methods rely on continuous hidden representations extracted from large-scale self-supervised learning (SSL) speech models such as wav2vec~2.0~\cite{baevski2020wav2vec}, HuBERT~\cite{hsu2021hubert}, and Whisper~\cite{radford2023whisper}, as input to temporal decoders. For instance, FaceFormer~\cite{fan2022faceformer} and CodeTalker~\cite{xing2023codetalker} introduced autoregressive transformers over wav2vec~2.0 features, while FaceDiffuser~\cite{stan2023facediffuser} reframed the task as a diffusion-based decoder over HuBERT representations. 
SSL representations are learned from massive amounts of speech data using a \textit{masking pretext task}, which consists of training the model to fill-in gaps in the input signal from the surrounding context using discrete signal representations. Therefore, those models inherently specialize in encoding discrete speech events, such as phonetic classes, but in a highly contextualized setting sometimes called \textit{``semantic representation''}. 
The success of such SSL representations in speech-driven animation suggests that highly contextualized segmental information provides sufficient cues for predicting realistic continuous articulatory trajectories to drive facial motions. 


In parallel, recent speech generation pipelines have shifted toward using neural audio codecs (e.g., WavTokenizer, BigCodec) \cite{ji2024wavtokenizer,Xin2024} to compress speech signals into a low-dimensional bottleneck. In this latent space, the resulting representations are less constrained and do not explicitly encode structured segmental or suprasegmental information \cite{Sadok2025b}. Consequently, these are commonly referred to as \textit{acoustic representations}.
In between, SpeechTokenizer~\cite{zhang2024speechtokenizer} uses successive residual vector quantized layers as speech representation, and combines a reconstruction pretext task with the distillation of HuBERT representations on the first layer to lead to a hybrid \textit{semantic+acoustic representation}.
Towards discrete speech information modelling, CosyVoice2~\cite{du2024cosyvoice2} trains a speech encoder with an \textit{Automatic Speech Recognition pretext task}, including labels for characters and prosodic events (e.g., emphasis, speaking rate). Therefore, this training may push the model to create representations that mostly encode discrete speech information. This representation space will be referred to as \textit{label-based} from now on.
While all these speech representations can be decoded into speech with various levels of quality, the same question arises for facial animation: 
\emph{which information should a speech representation expose to best drive 3D facial animation?} 


A second aspect is that most speech representations used for audio reconstruction are quantized (or tokenized). 
This change of paradigm 
may be one reason why, 
despite a growing interest in discrete speech representations, their suitability for speech-driven facial animation remains fairly unexplored. VQTalker maps discrete speech tokens to quantized facial motion tokens for multilingual 2D talking head synthesis, showing that discrete-to-discrete mappings can outperform continuous alternatives in that setting \cite{liu2025vqtalker}. SOLAMI employs discrete speech and motion tokens within a social interaction framework for 3D autonomous characters, focusing on body gestures \cite{jiang2025solami}. Interestingly, both methods use SpeechTokenizer as a speech encoder, which combines the semantic representation of HuBERT with an additional acoustic representation, but neither examines the relevance of using such hybrid representations for facial articulation prediction. 
Inversely, and to the best of our knowledge, no study has investigated the use of either fully \textit{acoustic} representations (no encoding of discrete phonetic classes) or fully \textit{label-based} representations (mostly encoding discrete events) to predict 3D facial animations.
Thus, it remains unclear whether token spaces capture phoneme-level classes, articulatory dynamics, or higher-level abstractions that can be efficiently mapped to 3D facial deformations.




In this work, we present a systematic evaluation of four speech representations for 3D facial information, including \textit{acoustic}, \textit{semantic+acoustic}, \textit{semantic}, and \textit{label-based}, with a focus on the information they make accessible for facial decoding (see Fig. \ref{fig:general_pipeline}). Our contributions are the following: (1)~We compare four speech encoders spanning the aforementioned speech representations across two temporal facial decoding architectures. Performance is assessed using objective metrics and a perceptual evaluation to compare animation realism and lip-speech coherence subjectively. (2)~In addition, we conduct probing analyses that relate speech representations to phonetic units, reflecting discrete class-like content, and to blendshape clusters, reflecting articulatory correlation. (3)~Finally, we observe that tokenized speech representations open a practical opportunity: since text-to-speech (TTS) systems already produce such tokens as an intermediate step (e.g.,~\cite{du2024cosyvoice2,wang2023valle}), a single representation can be decoded simultaneously into speech audio and facial motion, as done by~\cite{Lenglet2024} with continuous textual representations of a FastSpeech~\cite{Ren2021} model. We exploit this to introduce a proof-of-concept of an Audio Visual Text-to-Speech (AVTTS) pipeline that generates synchronized speech and 3D facial animation from text, without requiring a separate audio-driven stage.

\section{Method}

\subsection{Experimental Setup}
\label{sec:experimental-setup}

To investigate which speech tokens can serve as effective representations for speech-driven 3D facial animation, we adopted a comparative experimental framework.
The general pipeline of our method is illustrated in Fig. \ref{fig:general_pipeline}. 
We evaluated four speech encoders: HuBERT (\HuBERT)~\cite{hsu2021hubert}, SpeechTokenizer (\SpeechTokenizer)~\cite{zhang2024speechtokenizer}, WavTokenizer (\WavTokenizer)~\cite{ji2024wavtokenizer}, and CosyVoice2 (\CosyVoice)~\cite{du2024cosyvoice2}. Each encoder is paired with one of two facial decoders: a Gated Recurrent Unit (\GRU)~\cite{cho2014gru} or a Transformer (\T) architecture \cite{vaswani2017transformer}. [\HuBERT+\GRU] serves as our SSL representation baseline (\B) and is the exact reproduction of the FaceDiffuser~\cite{stan2023facediffuser} architecture and training\footnote{https://github.com/uuembodiedsocialai/FaceDiffuser}. The remaining three models are discrete tokenizers that cover different representation learning designs: SpeechTokenizer uses a multi-codebook with \textit{semantic} distillation; WavTokenizer employs a single-codebook with extreme compression aiming at \textit{acoustic} reconstruction; and CosyVoice2 relies on a supervised \textit{label-based} generative framework. 
Regarding the facial decoders, the GRU operates frame-by-frame, while the non-causal Transformer processes full sequences via self-attention. 
All combinations of encoder and decoder have been done relatively to the \B architecture to have minimal pair comparisons between encoder or decoder variations. Among all variants, [\HuBERT+\T], [\SpeechTokenizer+\GRU], and [\SpeechTokenizer+\T] combinations can be found in the literature even though under different architectures than \B~\cite{jiang2025solami,wu2024probtalk3d}. We call these variants \Vsota. The remaining combinations involving WavTokenizer and CosyVoice2 are, to the best of our knowledge, never been tested for facial animation prediction. We call these variants \Vour.

\begin{figure}[t!]
    \centering
    \includegraphics[width=1.0\linewidth]{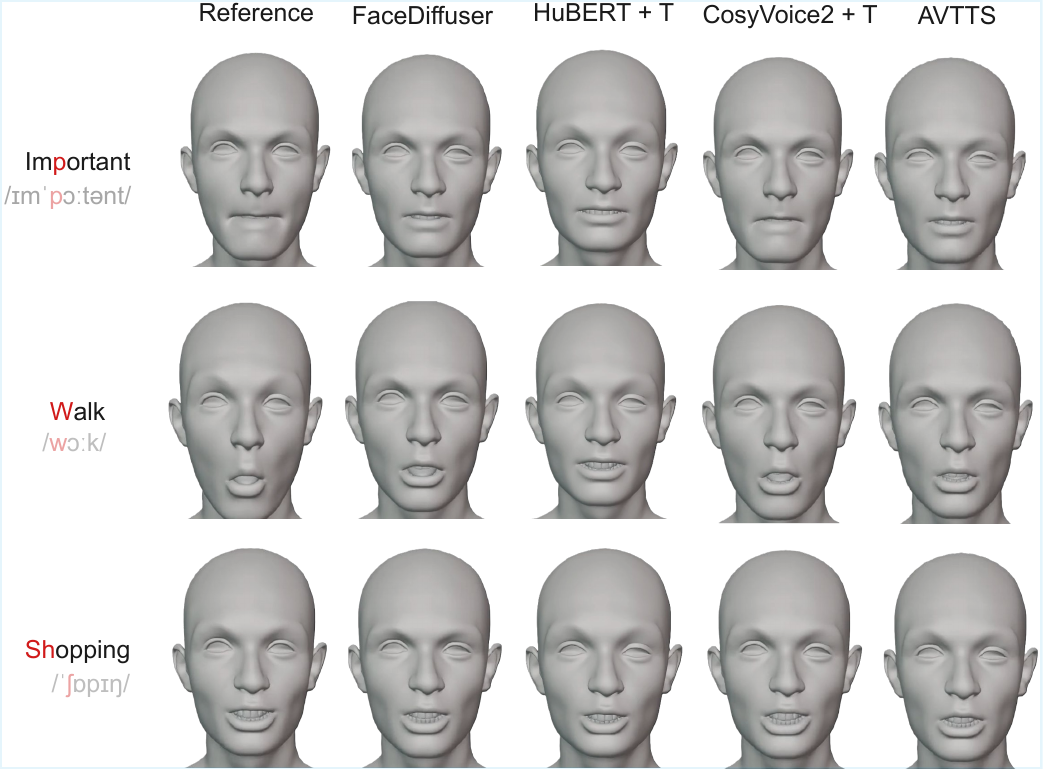}
    \caption{Sequence of rendered blendshapes across different models and audio snippets. With each different line, we are able to visually assess mouth and lip shapes across different groups of phonemes with significantly distinct visemes. Mesh model adapted from EmoTalk \cite{peng2023emotalk}.}
    \label{fig:visual_analysis}
    \vspace{-1.5em}
\end{figure}


Each encoder–decoder combination was trained with the encoder frozen using pre-trained weights, and the decoder trained from scratch on the BEAT2 dataset \cite{liu2024emage}.
This dataset provides approximately 27 hours of English speech waveforms from 25 different speakers (native and non-native) in scripted monologues, uttered in 8 different basic emotions. They are aligned with 3D facial motion in FLAME parameter space. To standardize the output, we convert these FLAME parameters to 51-dimensional ARKit blendshapes using a transformation matrix provided by the dataset authors~\cite{arkit2017}. The facial decoders map the extracted speech representations to these ARKit blendshapes. Inspired by \cite{stan2023facediffuser}, the GRU serves as the denoising network in a diffusion process, trained with $L_1$ reconstruction loss for direct data prediction instead of noise prediction. The Transformer consists of decoder layers with cross-attention, and was trained with $L_1$ loss combined with first- and second-order motion smoothness losses (velocity and acceleration) to encourage temporal coherence.

Fig.~\ref{fig:visual_analysis} shows a visual comparison of rendered sequences of blendshape frames. [\CosyVoice+\T] output is slightly closer to the reference when compared with FaceDiffuser (\B [\HuBERT+\GRU]), and in pair with the continuous speech representation approach of [\HuBERT+\T]. In the following section, we introduce quantitative metrics to assess these differences.

\subsection{Evaluation Metrics}


\noindent\textbf{Facial Animation Metrics.}
We evaluated reconstruction using Lips Vertex Error (LVE) adapted for blendshapes, which computes the mean squared distance between the predicted and ground-truth lip, jaw, and mouth blendshapes. However, recent studies \cite{haque2025wild, nocentini2024beyond} have shown that most reconstruction objective metrics often correlate poorly with human perception of synthetic facial animations. To address this, we additionally reported Jitter Score, which measures the second-degree derivative (acceleration) of blendshape frames, and Bilabial Closure Score (BCS), to better capture articulatory accuracy.
Our proposed BCS evaluates effective lip closure during bilabials \textipa{/b/, /p/} and \textipa{/m/}: we first calculated on the ground truth data a 
threshold which bounds the blendshape values on $70\%$ of the bilabial frames, i.e., which defines the lip closure position.
Second, we calculated the number of predicted frames that passed this threshold on bilabial phones and divided it by the total number of bilabial frames to obtain the BCS. 
The impact of the speech encoder and the facial decoder on LVE and Jitter was investigated using linear regression models (R function \texttt{lme}). Post-hoc pairwise comparisons were conducted using
the \texttt{emmeans} package, with significance set at $p < 0.05$.

\noindent\textbf{Perceptual Evaluation.} 
From all encoder-decoder combinations, we choose three meaningful representatives: [\HuBERT+\GRU], our \B model; [\HuBERT+\T], a \Vsota variant 
to assess decoder influence compared to \B; and [\CosyVoice+\T], the \Vour variant which obtained the best objective scores (section~\ref{sec:results}). 
The evaluation was conducted on Prolific following a MUSHRA-like protocol: for a given utterance, participants were asked first to watch a reference video, and then rate, from 0 to 100, four videos (the three systems and a hidden reference identical to the presented reference video) of rendered blendshapes based on how close the facial animation appears compared to the reference. Participants rated 15 stimuli randomly selected from the test set.
To promote data reliability and ensure adequate task engagement, participants whose average rating of the hidden reference was below 80\% were excluded from further analysis. While lower scores on the hidden reference may reflect differences in scale usage or individual rating strategies, they can also indicate reduced task engagement or difficulty in recognizing stimulus equivalence within the MUSHRA framework. After applying this criterion, the final sample consisted of 30 participants.
We assessed the effect of the model on the MUSHRA scores using a beta regression model (R function \texttt{glmmTMB}), followed by post-hoc pairwise comparisons (R function \texttt{emmeans}), with significance set at $p < 0.05$.

\noindent\textbf{Probing Metrics.}
To better understand how the variety of speech representations under study encode speech and facial information, we conducted probing of phonetic classes and facial information for all four encoders. For consistency, we also quantized HuBERT representations.
We first aligned phonetic and facial information with speech tokens of all models for each frame. We used the phone alignment provided with the dataset to assign a phonetic class to each token.
For facial features, we focused on both discrete representations in the form of canonical facial poses, which we call \textit{visemes}, and continuous representations in the form of raw blendshapes prediction. To obtain the visemes, we clustered ground-truth blendshapes vectors from the test set using the $k$-means algorithm, discretizing the 51D continuous representations into 32 visemes.
Then, we aligned predicted tokens with facial motion captured at 30 FPS using nearest-neighbour temporal matching. We finally associated the nearest \textit{viseme} to each token.

To probe both phonetic classes and visemes (i.e., discrete representations of speech and facial movements), we computed co-occurrence statistics between token IDs and phoneme or viseme labels across the test set, using normalized entropy:
%
%
\begin{equation}
    \hat{H}(\mathcal{X} \mid t) = \left[-\sum_{i=1}^{K} p(x_i \mid t) \log_2 p(x_i \mid t) \right] / \log_2{(K)}
\end{equation}
where $\mathcal{X}$ is either the set of $K$ visemes $\mathcal{V}$ or phonemes $\mathcal{P}$. $p(x_i \mid t)$ is the fraction of times token $t$ co-occurs with target $x_i \in \mathcal{X}$. The term $\log_2({K})$ corresponds to the maximum entropy, i.e., when predicting one sample among $K$ with uniform distribution (chance level). Normalising by this term bounds $\hat{H}$ between 0 (minimum entropy, one-to-one mapping between all tokens and visemes/phonemes) and 1 (maximum entropy, chance level to predict all visemes/phonemes from any token).

To complement this probing of discrete phoneme and viseme representations, we also employ a Ridge regression probe to directly predict continuous blendshape values from sequences of one-hot token features. We report the median coefficient of determination $R^2$ across all blendshapes as a measure of a (linear) continuous representation of facial movements within the speech representation space.

\begin{table*}[htbp]
\centering
\caption{Objective metrics from predicted blendshape sequences and probing analysis on speech representations. \B is FaceDiffuser~\cite{stan2023facediffuser}, \Vsota are variants of \B found in SOTA, \Vour are novel variants. Pairs of symbols indicate when the difference between the two reported metrics is not statistically significant ($p < 0.05$). Bold (resp. underline) indicates best (resp. second best) scores.}
\label{tab:general_results}
\begin{tabular}{@{}cclcccccc@{}}
\toprule
\textbf{Speech representation} & \textbf{Facial decoder} & \textbf{Type} & \textbf{LVE $\downarrow$} & \textbf{Jitter $\downarrow$} & \textbf{BCS (\%) $\uparrow$} & \textbf{\Hp\xspace(\%) $\downarrow$} & \textbf{\Hv\xspace(\%) $\downarrow$} & \textbf{\R\xspace$\uparrow$} \\
\midrule
\midrule
\multirow{2}*{$\begin{matrix} \textit{semantic} \\ \text{\scriptsize{(HuBERT \cite{hsu2021hubert})}} \end{matrix}$} 
  & GRU   & \B & \textbf{0.26} \symbsigdiamond & 80.3 \symbsigblank          & \textbf{57.5}            & \multirow{2}*{\underline{44.4}}    & \multirow{2}*{91.5}    & \multirow{2}*{\textbf{0.25}}\\
  \cmidrule{2-6}
  & Trans. & \Vsota & \textbf{0.26} \symbsigdiamond          & \underline{45.5} \symbsigcircle & 27.6            &    &      & \\
\midrule
\multirow{2}*{$\begin{matrix} \textit{semantic+acoustic} \\ \text{\scriptsize{(SpeechTokenizer \cite{zhang2024speechtokenizer})}} \end{matrix}$}
  & GRU    & \Vsota & 0.53 \symbsigblank          & 74.4 \symbsigsquare          &  3.4          & \multirow{2}*{\textbf{39.6}}  & \multirow{2}*{90.4}  & \multirow{2}*{0.04}\\
  \cmidrule{2-6}
  & Trans. & \Vsota & 0.34 \symbsigblank          & \textbf{35.2}  \symbsigblank &  2.3 &  &   & \\
\midrule
\midrule
\multirow{2}*{$\begin{matrix} \textit{acoustic} \\ \text{\scriptsize{(WavTokenizer \cite{ji2024wavtokenizer})}} \end{matrix}$}
  & GRU    & \Vour & 0.84 \symbsigblank          & 93.3  \symbsigblank          & 0.4            & \multirow{2}*{73.2} & \multirow{2}*{91.4} & \multirow{2}*{0.08} \\
  \cmidrule{2-6}
  & Trans. & \Vour & 0.33 \symbsigblank          & \underline{43.6} \symbsigcircle          & 6.2            &  &  & \\
\midrule
\multirow{2}*{$\begin{matrix} \textit{label-based} \\ \text{\scriptsize{(CosyVoice2 \cite{du2024cosyvoice2})}} \end{matrix}$}
  & GRU    & \Vour & 0.46 \symbsigblank          & 76.5 \symbsigsquare          & 3.9            & \multirow{2}*{57.7} & \multirow{2}*{\textbf{84.3}}
 & \multirow{2}*{0.10}\\
  \cmidrule{2-6}
  & Trans. & \Vour & \underline{0.28} \symbsigblank & 50.3 \symbsigblank          & \underline{47.0}            &  & 
 & \\
\bottomrule
\end{tabular}%
\end{table*}

\section{Results and Discussion}
\label{sec:results}

\noindent\textbf{Objective metrics.} 
Table \ref{tab:general_results} shows the results for the objective metrics on generated animation sequences by our 8 model variants on the BEAT2 test set (265 stimuli, or approximately 4 hours). In terms of lips and mouth accuracy (LVE), both HuBERT models score higher than the discrete models, although the performance of [\CosyVoice+\T] is not far behind. Also, models with discrete speech representations perform significantly better when coupled with the Transformer architecture, while the continuous HuBERT representation shows equivalent results between decoders. For Jitter, all speech representations decoded with a Transformer get better results than those with a GRU. In terms of lip closure (BCS), the \B model gets the highest score, closely followed by [\CosyVoice+\T], then [\HuBERT+\T]. All other models display very low values, showing an absence of expected lip closure during bilabial phonemes.

\begin{figure}[t!]
    \centering
    \includegraphics[width=1.0\linewidth]{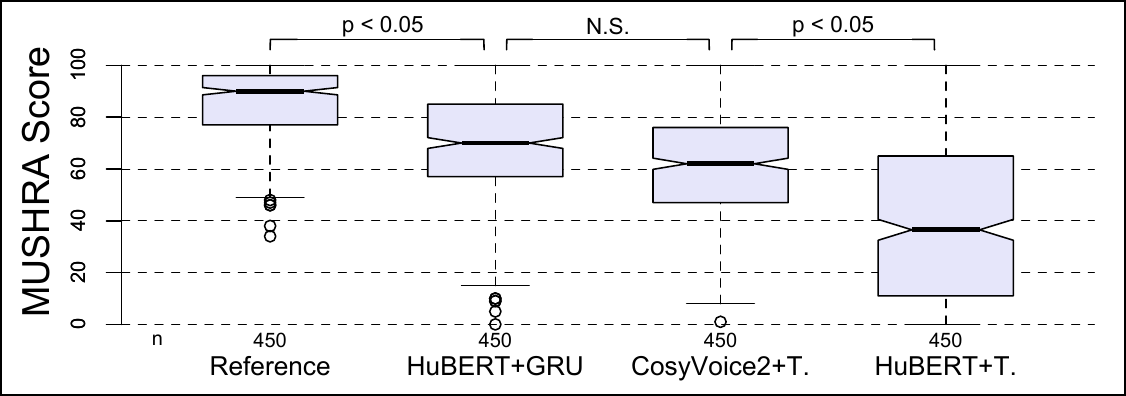}
    \caption{MUSHRA-like scores of the perceptual evaluation across three different models and the reference. N.S. means no statistical significance between the pair of distributions.}
    \label{fig:mushra_results}
    \vspace{-1.5em}
\end{figure}

\noindent\textbf{Perceptual Evaluation.} Fig. \ref{fig:mushra_results} shows the results from the perceptual evaluation. 
We observe three groups of conditions: the reference rated significantly higher, followed by [\HuBERT+\GRU] (\B) and [\CosyVoice+\T] (\Vour) with a non-significant difference between the two, and finally [\HuBERT+\T] (\Vsota).
Compared with Table~\ref{tab:general_results}, we can observe that the evaluated setups in the perceptual evaluation follow the same order of performance on BCS, which is not the case for LVE. This indicates, for this work, a closer correlation between perceptual analysis and lip-closing metrics, instead of traditional reconstruction metrics.

\noindent\textbf{Probing speech representations.} 
The last three columns of Table \ref{tab:general_results} show the results for the probing stage on the speech representations. 
The \textit{semantic+acoustic} representation provides the best representation of phonetic classes (\Hp), followed by \textit{semantic}, \textit{label-based}, and finally \textit{acoustic}. This shows that distilling HuBERT within SpeechTokenizer is even more efficient than HuBERT representations themselves to encode phonetic information. As for \textit{label-based} representations, the training task is character-oriented (as opposed to phoneme-oriented), and could explain the lower scores. The low score of \textit{acoustic} representations is consistent with previous observations~\cite{Sadok2025b}. Globally, if all models encode little information about visemes, the \textit{label-based} representation stands out, which could have been favoured by a fully supervised discrete output training task.
Finally, compared to HuBERT, all three tokenized representations display a very low $R^2$ metric, which measures how accurately a blendshape trajectory can be predicted from tokens, which is intrinsically consistent with the continuous/discrete nature of the representations.

\noindent\textbf{Discussion.} 
By examining evaluations, the first striking result is the adequacy of the \textit{semantic} representation (\B) for facial animation prediction, both in terms of objective and perceptual metrics. This justifies that \textit{semantic} representations that encode well phonetic classes are a suitable representation for the task. 
Very interestingly, our proposed variant [\CosyVoice+\T] is very competitive in terms of performance: it is not significantly different than \B from a perceptual point of view, and its \textit{label-based} speech representation also allows the encoding of phonetic classes to a certain extent. 
Surprisingly, hybrid \textit{semantic+acoustic} representations did not provide satisfactory facial animation prediction, despite their excellent encoding of phonetic information. Results are comparable with \textit{acoustic} representations only, suggesting that the presence of low-structured \textit{acoustic} representations might be detrimental to the prediction of facial animation.
Overall, these results suggest that phonetic class information could be a necessary but not sufficient condition for accurate facial animation prediction, as it should be accompanied by an absence of low-structured acoustic information.
As for the necessity of encoding facial representation, we can note that the best predicted models, \B and [\CosyVoice+\T], stand out at encoding continuous blendshape and discrete visemes, respectively. Nevertheless, these values remain low, and further investigation is needed to evaluate the necessity of those representations for accurate prediction.

\section{Towards a Unified AVTTS Pipeline}
\label{sec:all-pipeline}
To further explore the capabilities of speech tokens to be used for facial animation, we propose a \emph{proof-of-concept} of a shared representation for both speech and face synthesis in a unified framework (see Fig. \ref{fig:general_pipeline}). 
The CosyVoice2 TTS model predicts speech tokens from text and a reference audio using an autoregressive LLM backbone. Thus, we simply employ our facial Transformer 
decoder, which was already trained on CosyVoice2 representations, to predict blendshape sequences.
This is done in parallel to the flow-matching speech decoder, which generates the speech waveform.
Since both decoders operate from the same token sequence, the resulting speech and facial motion are inherently synchronized, while it also eliminates the conventional two-stage paradigm in which text is first synthesized into audio and then processed by a separate speech-driven animation model. 
Finally, if Fig. \ref{fig:visual_analysis} displays less contrast of lip movement across phonemes with AVTTS compared with the audio-driven version [\CosyVoice+\T], our demo page\footnote{https://github.com/ProdCor/Token-to-Face}
displays encouraging results, which open the door to dedicated work on AVTTS using discrete representations.



\section{Conclusion}

This work presented a systematic comparison of \textit{semantic}, \textit{semantic+acoustic}, \textit{acoustic}, and \textit{label-based} speech representations for 3D facial animation generation.
Our results demonstrated that both \textit{semantic} and \textit{label-based} representations are suitable candidates for this task, and that they have similar perceptual performance. Probing analysis revealed that phonetic information encoding seems to be a necessary condition for the task, provided that low-structured acoustic information is not present. We also found a correlation between high performance and the encoding of either discrete or continuous facial representations. Finally, we showed that the shared representation between speech synthesis and facial animation enables a unified Audio Visual Text-to-Speech pipeline, which is an interesting alternative to the conventional two-stage generation paradigm. More broadly, this paves the way to more versatile multimodal speech language models, in which decoders can be easily adapted to frozen LLM backbones.

\section{Acknowledgments}
This work was funded by the São Paulo Research Foundation (FAPESP) under grant \#2025/09875-7 through a Research Internship Abroad (BEPE) scholarship in the GIPSA-lab (Université Grenoble Alpes), supported by FAPESP under grant \#2020/09838-0 (BI0S - Brazilian Institute of Data Science), and partially funded by the Coordenação de Aperfeiçoamento de Pessoal de Nível Superior – Brasil (CAPES) – Finance Code 001. The first author is affiliated with the Artificial Intelligence Lab, Recod.ai, and by MIAI Cluster (ANR-23-IACL-0006).

\section{Generative AI Use Disclosure}
The writing of this paper was supported by generative AI tools, which were used strictly for the refinement of the text to follow correct English grammar, sentence structure, and clarity.


\bibliographystyle{IEEEtran}
\bibliography{mybib}

\end{document}